\documentclass[12pt]{article}
\usepackage{times}

\usepackage{amsmath}
\usepackage{amssymb}
\usepackage{float}
\usepackage{graphicx}

\usepackage[comma,sort&compress]{natbib}
\usepackage{color} 

\usepackage{lineno}
\usepackage{url}
\usepackage{pdfpages}
\usepackage{setspace} 
\usepackage{lineno}

\usepackage{hyperref}

\hypersetup{
	bookmarks=true,         % show bookmarks bar?
	unicode=false,          % non-Latin characters in Acrobat‚????s bookmarks
	pdftoolbar=true,        % show Acrobat‚????s toolbar?
	pdfmenubar=true,        % show Acrobat‚????s menu?
	pdffitwindow=false,     % window fit to page when opened
	pdfstartview={FitH},    % fits the width of the page to the window
	pdftitle={My title},    % title
	pdfauthor={Author},     % author
	pdfsubject={Subject},   % subject of the document
	pdfcreator={Creator},   % creator of the document
	pdfproducer={Producer}, % producer of the document
	pdfkeywords={keywords}, % list of keywords
	pdfnewwindow=true,      % links in new window
	colorlinks=true,       % false: boxed links; true: colored links
	linkcolor=blue,          % color of internal links
	citecolor=blue,        % color of links to bibliography
	filecolor=magenta,      % color of file links
	urlcolor=cyan           % color of external links
}

\usepackage[labelfont=bf,labelsep=period,justification=raggedright]{caption}

\makeatletter
\renewcommand{\@biblabel}[1]{\quad#1.}
\makeatother

\date{}

\title{\bf{Voluntary safety commitments provide an escape from over-regulation in AI development} }
\author{The Anh Han$^{1,\star}$, Tom Lenaerts$^{2,3}$, Francisco C. Santos$^{2,4}$, and Lu\'is Moniz Pereira$^{5}$}

\begin{document}
	\maketitle
	{\footnotesize
		\noindent
		$^{1}$ School of Computing, Engineering and Digital Technologies,  Teesside University, Middlesbrough, UK TS1 3BA\\
		$^2$ Machine Learning Group, Universit{\'e} Libre de Bruxelles, Boulevard du Triomphe CP212, 1050 Brussels, Belgium  \\ 
		 $^3$ Artificial Intelligence Lab, Vrije Universiteit Brussel, Boulevard de la Plaine 2, 1050 Brussels, Belgium
		 	$^{4}$	INESC-ID and Instituto Superior Tecnico, Universidade de Lisboa, Portugal \\
		 $^5$ NOVA Laboratory for Computer Science and Informatics (NOVA LINCS), %Faculdade de Ci\^encias e Tecnologia, 
		Universidade Nova de Lisboa, 
		2829-516 Caparica, Portugal \\
		\\ \\
		$^\star$ Corresponding author: The Anh Han (theanhhan.vn@gmail.com) 
	}

\newpage
\section*{Abstract}
With the introduction of Artificial Intelligence (AI) and related technologies in our daily lives, fear and anxiety about their misuse as well as the hidden biases in their creation have led to a demand for regulation to address such issues.  Yet blindly regulating an innovation process that is not well understood, may stifle this process and reduce benefits that society may gain from the generated technology, even under the best intentions. In this paper, starting from a baseline model that captures the fundamental dynamics of a race for domain supremacy using AI technology, we demonstrate how socially unwanted outcomes may be produced when sanctioning is applied unconditionally to risk-taking, i.e. potentially unsafe, behaviours. As an alternative to resolve the detrimental effect of over-regulation, we propose a voluntary commitment approach wherein technologists have the freedom of choice between independently pursuing their course of actions or establishing binding agreements to act safely,  with sanctioning of those that do not abide to what they pledged. Overall, this work reveals for the first time how voluntary commitments, with sanctions either by peers or an institution, leads to socially beneficial outcomes in all scenarios envisageable in a short-term race towards domain supremacy through AI technology.  These results are directly relevant for the design of governance and regulatory policies that aim to ensure an ethical and responsible AI technology development process. \\
 
 \noindent \textbf{Keywords:} Evolutionary Game Theory, AI development race, Commitments, Incentives, Safety. % \ty{dynamics of innovation and cooperation}.

 \newpage
 
\doublespacing

\section{Introduction}
With the rapid advancement of  AI and related technologies, there has been significant fear and anxiety about their potential misuse as well as the social and ethical consequences that may result from biases within the design of such systems \citep{tzachor2020artificial,bostrum2014superintelligence,stix2020bridging}. 
While expectations associated with these advanced technologies increase and monetary profits stimulate rapid deployment, there is a serious risk for taking unethical or risky short cuts to enter a market first with the next innovation, ignoring safety checks and ethical development procedures.   
As different disagreeable examples have emerged \citep{coeckelbergh2020ai}, governments and regulating bodies have been catching up by debating new forms and frameworks for regulating this technology     \citep{baum2017promotion,cave2018ai,taddeo2018regulate}; notably, the recent EU White Paper on AI.    
Such debates have produced proposals for mechanisms on how to avoid, mediate, or regulate the development and deployment of  AI  \citep{baum2017promotion,cave2018ai,geist2016s,shulman2009arms,hanAIES2019,vinuesa2019role,nemitz2018constitutional,taddeo2018regulate,askell2019role,o2020windfall}. Essentially, regulatory measures such as restrictions and incentives are proposed to limit harmful and risky practices in order to promote beneficial designs \citep{baum2017promotion}.  Examples of such approaches \citep{baum2017promotion} include financially supporting the research into beneficial AI \citep{mcginnis2010accelerating} and  making AI companies pay
fines when found liable for the consequences of harmful AI \citep{gurney2013sue}.  

Although these regulatory measures may provide solutions for particular scenarios, one needs to ensure that they do not overshoot their targets since over-regulation could stifle innovation,  potentially hindering investments into the development of novel innovations as they become too risky endeavors \citep{hadfield2017rules,lee2018ai}.   
Worries  have been expressed by different organisations/societies that too strict policies may unnecessarily affect the benefits and societal advances that novel AI technologies may have to offer \citep{ERDI2021}.  
Regulations affect moreover big and small tech companies differently: A highly regulated domain makes it more difficult for small new start-ups, introducing an inequality and dominance of the market by a few big players \citep{lee2018ai}. It has been emphasised that neither over-regulation nor a laissez-faire approach suffices when aiming to regulate AI technologies \citep{Australia2019}. In order to find a balanced answer, one clearly needs to have first an understanding of how a competitive development dynamic may work.

Starting  from a baseline game theoretical model, referred to as the DSAIR model \citep{han2019modelling} (a model of domain supremacy through an AI race), which defines the process through which multiple stake-holders aim for market supremacy,  we demonstrate first that  unconditional sanctioning will negatively influence social welfare in certain conditions of a short-term race towards domain supremacy through AI technology.  Afterwards, we examine the DSAIR model with an alternative mechanism for resolving the issue that can be shown to lead to less detrimental effects. Our approach is to allow technologists or race participants to voluntarily commit themselves to safe innovation procedures, signaling to others their intentions. Specifically, this  bottom-up, binding  agreement (or commitment) is established for those who want to take a safe choice,  with sanctioning applied to violators of such an agreement. 

\paragraph{Previous Developments.}
As was shown in \citep{han2019modelling}  participants either follow safety precautions (the SAFE option) or ignore them (the UNSAFE option) in each step of the development process of the DSAIR model. The main assumption in the model was that it requires more time and more effort to comply with the precautionary requirements, making the SAFE option not only costlier, but  also slower compared to the UNSAFE option.  Accordingly, it was assumed that in playing SAFE, participants must pay a cost $c > 0$, whereas playing UNSAFE costs nothing.  Furthermore, whenever playing UNSAFE, the development speed differs and is $s > 1$  whereas in playing SAFE the speed is simply normalised to 1.
Decisions to act SAFE or UNSAFE in AI development are repeated until one or more teams attain the designated objective, which can be translated into having completed $W$ development steps, on average \citep{han2019modelling}. As a result, they earn a large benefit or prize $B$ (e.g. windfall profits \citep{o2020windfall}), equally shared among those  reaching the target at the same time. A development disaster or a setback  may however come to occur with some probability, which is presumed to increase with the number of times that safety requirements were ignored by the winning team(s) at each step. 
 Whenever a disaster of this kind occurs, all the benefits of a risk-taking participant are lost.  This risk probability is denoted by $p_r$ (see the Models and Methods section for more details).

\begin{figure}
\centering
\includegraphics[width=0.5\linewidth]{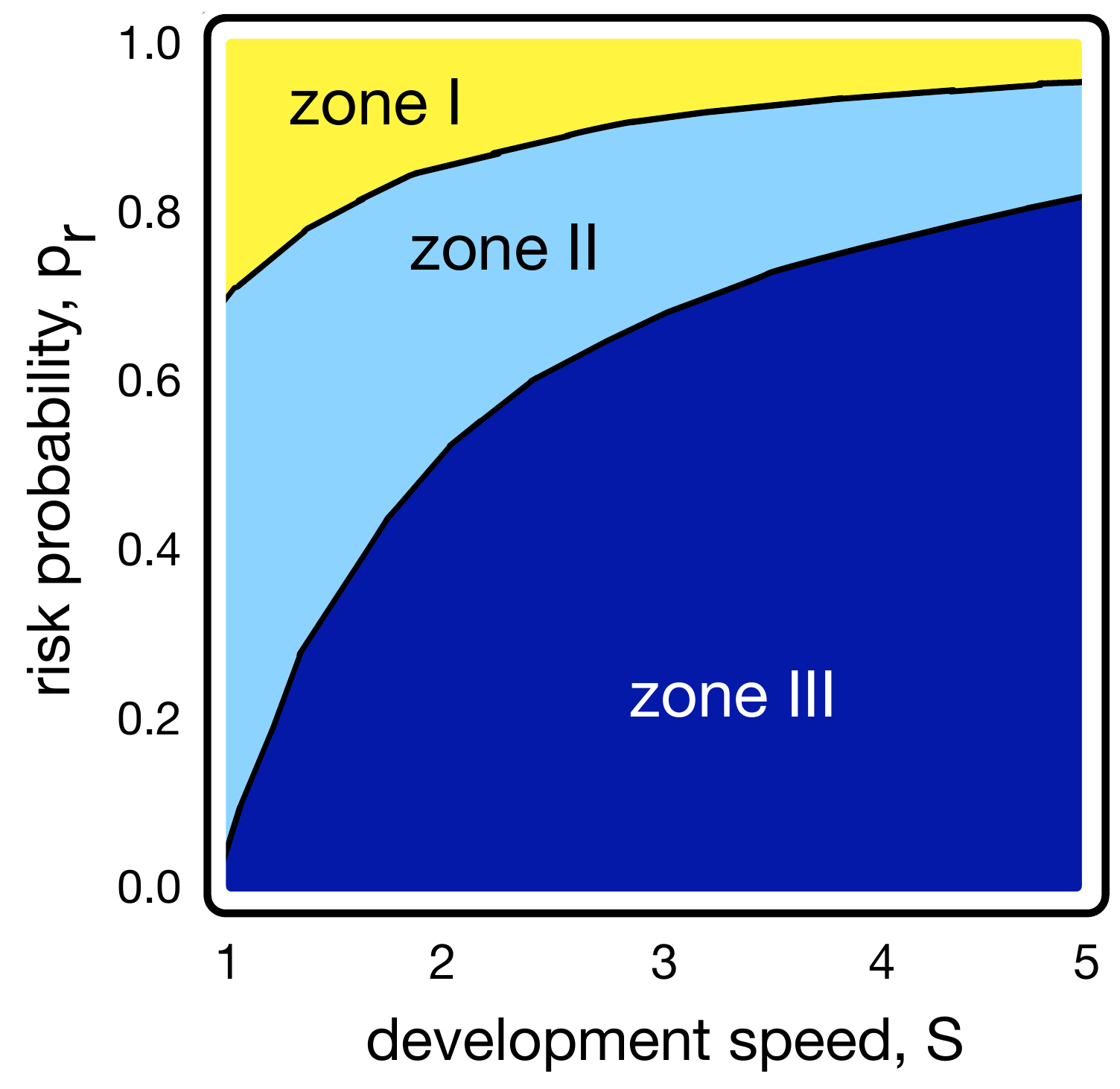} 
\caption{\textbf{Behavioural regions (zones) as identified in \citep{han2019modelling} when the time to reach domain supremacy is short.} Region \textbf{II}: Inside the plots, the two solid lines  delimit the boundaries wherein the collective prefers safety compliant behavior, yet unsafe development is individually preferred. Regions \textbf{III} and \textbf{I} exhibit where unsafe (respectively, safe) development is both the individually and collectively preferred outcome}
\label{fig:panel_no_punishment}
\end{figure}

It was observed in the DSAIR model that, in case the time-scale to reach the target is short, so that over the whole of the development process the average of the accumulated benefit is much smaller than the  final benefit $B$, only for a certain window of parameter settings societal interest conflicts with the individual ones: In that region, individual unsafe behaviour dominates, despite that safe development would lead to a larger collective outcome or social welfare (cf. region \textbf{II} in Figure \ref{fig:panel_no_punishment}). 
From the regulatory perspective, it is only region \textbf{II} that thus requires governance in order to promote or enforce safe actions in order to avoid any disaster that may occur during the technology development race. 

A peer punishment mechanism against unsafe behaviour was proposed in order to mediate the behavior in that region (without affecting the desirable safe outcome in region \textbf{I}) \citep{han2020Incentive}. It however may lead to a reduction of the societal welfare that can be obtained in region \textbf{III} (see Figure \ref{fig:panel_no_punishment}) where the desired unsafe (risk-taking/innovative) behaviour becomes significantly reduced whenever punishment is not very costly to the punisher while strongly affecting the punished.  The problem with applying system-wide sanctioning for any risk-taking behavior is that it does not take into account the region wherein the AI innovation is taking place, as was visualised by Figure \ref{fig:panel_no_punishment}. That is, if it is the case that a development race falls in region \textbf{III} (low risk and innovation is collectively beneficial and preferred), unsafe behaviour should not be punished as it is beneficial for the overall social welfare. To enact such a region-dependent targeted punishment scheme, one would require the ability to estimate exactly the risk level associated with each AI development scenario, i.e. knowing beforehand the risk as well as speed of development. Clearly that may not  be easy due to the lack of data on how such an AI innovation dynamics works.  This is especially true  in the early stages of the development and adoption of many new technologies, which has become known as the so-called  double-bind problem \citep{Collingridge1980}: the impact of technologies including  AI ones  cannot be predicted until it becomes a reality, and controlling or changing it is difficult or no longer possible at that point. 

\paragraph{Objectives}
In this article an alternative solution is proposed that circumvents the need of being able to estimate correctly the speed of development and risk in order to appropriately regulate UNSAFE behavior: race participants can choose whether or not to establish a bilateral commitment to act safely, which also exposes them to a sanction in case they do not uphold their commitment.  While, on the one had, the development teams are allowed to work in an UNSAFE manner without repercussions if they do not commit; a prior agreement allows, on the other hand, the safety compliant participants to identify easily unsafe ones while having also the capacity to punish the dishonest ones (who might also be sanctioned by an external party such as a   regulating  institution). 

Our results reveal that this freedom of choice, if enabled through a prior bilateral commitment to SAFE actions, can, on the one hand, avoid  over-regulating unsafe behaviour and, on the other, improve significantly the desired safety outcome in the dilemma zone when compared to punishment alone. Interestingly, this type of binding pledges has been argued to be of relevance within other types of global conflicts and dilemmas, such as  environmental governance \citep{barrett2003environment,cherry2013enforcing}.

\section{Models and Methods}
We first recall the DSAIR model with pairwise interactions then extend it with the option to bilaterally commit to acting safely and associated punishment of violations of such commitments.  
The Evolutionary Game Theory (EGT) methods being used to analyse the models will then be described. 
\label{section:models and methods}
%\ty{We first recall the innovation race or domain supremacy through AI race  (DSAIR) model as developed in \citep{han2019modelling}, before extending it with punishment and rewards. This model has been framed within the context of a race for technological supremacy through AI, but is general enough to be applicable to other innovation and technological races. }
\subsection{Summary of the DSAIR model and prior results}
The DSAIR model \citep{han2019modelling} was originally defined as a two-player game repeated with a certain probability, consisting thus on average of $W$ rounds\footnote{An N-player version of this game was discussed also in \citep{han2019modelling}.  Yet in order to keep things easy to access, we focus here on the two-player scenario.}. 
At each round of development,  players gather benefits arising from their intermediate AI developments, subject to whether or not they chose to act UNSAFE or SAFE. Presuming some fixed benefit, $b$, resulting from the AI market, the teams will share this gain proportionately to their development speed. 
Accordingly, at every round of the race one can write a payoff matrix denoted by $\Pi$ with respect to row players $i$, whose entries are denoted by $\Pi_{ij}$ ($j$ corresponding to some column),  as shown  
{\begin{equation}
\Pi =  \bordermatrix{~ & \textit{SAFE} & \textit{UNSAFE}\cr
                  \textit{SAFE} & -c + \frac{b}{2} &-c +  \frac{b}{s+1}      \cr
                  \textit{UNSAFE} &  \frac{s b}{s+1}   & \frac{b}{2}   \cr
                 }.
\end{equation}
}
The payoff matrix can be explained as follows. Firstly, wherever there is an interaction between two players selecting the SAFE action, each shall pay a cost $c$ and the resulting benefit $b$ is shared. Differently, whenever interaction is between two players selecting the UNSAFE action, they shall share benefit $b$ without having had to pay cost $c$. 
{Whenever an UNSAFE choice is matched with a SAFE one, the SAFE choice necessitates a cost $c$} {and receives a (smaller) part $b/(s+1)$ of  $b$, whereas the UNSAFE choice collects a larger  $sb/(s+1)$ whilst not ever having had to pay $c$}.  Note that $\Pi$ is a simplification of the matrix defined in \citep{han2019modelling} for, in the current time-scale, it was  shown that the parameters as defined here sufficiently explain the obtained results.

 We analyse the evolutionary outcomes of this game in a well-mixed and finite population consisting of $Z$ players. Given the choices each player can make and the fact that these choices need to be repeated for $W$ round, each adopts player one of the two following strategies \citep{han2019modelling}: 
\begin{itemize} 
	\item \textbf{AS}: complies every time with safety precautions, acting thus SAFE in every round.  
	\item \textbf{AU}: complies not once with safety precautions, acting thus UNSAFE in every round. 
\end{itemize} 
The averaged payoffs for {AS vs AU} are expressed by payoff matrix  
\begin{equation}
 \bordermatrix{~ & \textit{AS} & \textit{AU} \cr
                  \textit{AS} & \frac{B}{2W} +\Pi_{11} & \Pi_{12}    \cr
                  \textit{AU} &  p \left(\frac{sB}{W} + \Pi_{21}\right)   &  p \left(\frac{sB}{2W} +\Pi_{22}\right)  \cr 
              %    \textit{CS} &  \frac{B}{2W} +\Pi_{11}    & \frac{s}{W}  \left( \Pi_{12} + (\frac{W}{s} - 1) \Pi_{22}  \right) &  \frac{B}{2W} +\Pi_{11}    \cr
                 },
\end{equation}
wherein,  for presentation purposes alone, let us denote $p = 1 - p_r$ (note that $p_r$ was explained in the Introduction section).

As has been shown in \citep{han2019modelling}, by contemplating where AU is risk-dominant against AS (cf. Methods below), then three distinct regions are identifiable within the parameter space $s$-$p_r$ (cf. Figure \ref{fig:panel_no_punishment}): (\textbf{I})  if $ p_r > 1-\frac{1}{3s}$, AU is risk-dominated by  AS:   safety {compliance} affords both the collectively preferred  outcome and the one evolution selects; (\textbf{II}) if $1-\frac{1}{3s} > p_r > 1-\frac{1}{s}$:  even though  safety {compliance} is the more desirable strategy for ensuring the highest collective outcome,  the  {social learning dynamics}     {leads}  the population to the state {within which safety precautions} have been mostly ignored;  (\textbf{III}) if $p_r <  1-\frac{1}{s}$ (AU is risk-dominant against  AS),  then unsafe development is both {collectively preferred} and selected by the {social learning dynamics}.

So it is important to remember for the rest of the paper that UNSAFE actions are preferred and established in zone \textbf{III}, SAFE actions are preferred and established in zone \textbf{I} and a conflict exists between the individual and the collective in zone \textbf{II} as the former prefers UNSAFE actions and the latter SAFE actions.  %Numerical results in Figure \ref{fig:panel_no_punishment} confirm this division of the regions. 

\begin{figure*}[h!]
\centering
\includegraphics[width=\linewidth]{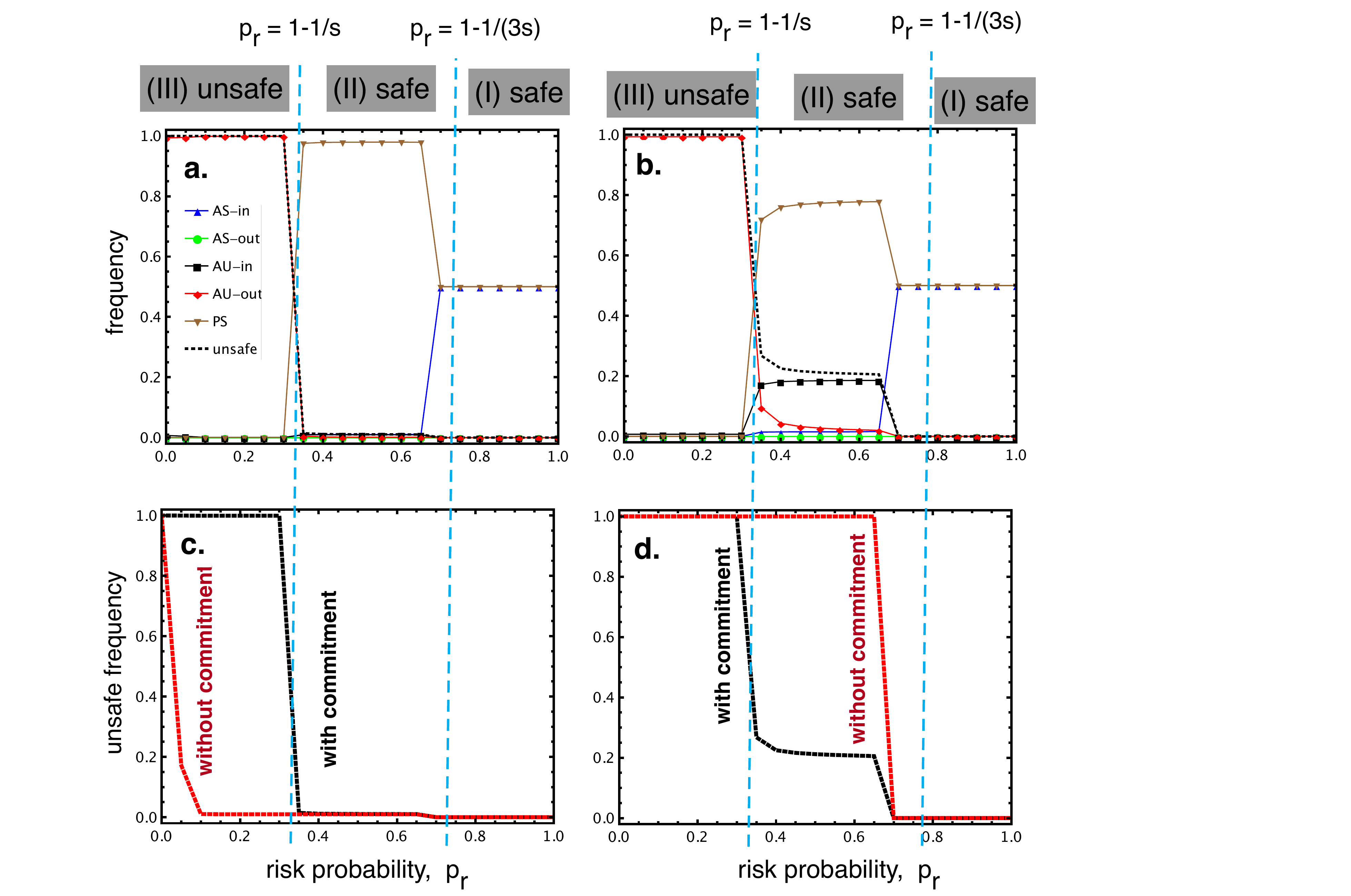} 
\caption{\textbf{Behavioural dominance in different zones for varying $p_r$, in presence of prior commitments  (top row: panels a, b) and comparison of its overall unsafe behaviour against when there are no commitments (bottom row: panels c, d)}. The black dotted lines in panels a and b indicate the total unsafe frequency (i.e. the sum of AU-in and AU-out frequencies). The desired collective behaviour is indicated for each zone (i.e. unsafe in zone \textbf{III} and safe in zones \textbf{I} and \textbf{III}). We show results for  two important scenarios: when efficient punishment can be made for a small cost (\textit{left column:} $s_\alpha = 0.3$, $s_\beta = 1$) and when punishment is not highly efficient  (\textit{right column:}, $s_\alpha = 1$, $s_\beta = 1$).   
%In both cases, AU-out dominates when $p_r$ is small (zone III), PS dominates when it is intermediate (first part of zone II),  while AS-out dominates when it is large (part of zone II and zone I). That results in the fact that when agreement is present desirable outcomes are achieved in all three zones: unsafe/innovation in (III) and safe in (I) and (II) (see bottom row), although a more efficient punishment leads to better safety outcome in zone II (compared panels c and d). In absence of agreement, 
%over-regulation  occurs (i.e. safe is abundant but not desired) for a large part of zone (I) when efficient punishment can be done for a small cost (panel c), while safety dilemma occurs (i.e. safe is desired but infrequent) for a large part of zone (II) when punishment is not highly efficient.   
Parameters:  $b = 4$, $c = 1$, $s = 1.5$, $W = 100$, $B = 10^4$, $\beta = 1$,  $Z = 100$. 
}
\label{fig:agreementvsnoagreement}
\end{figure*}

\subsection*{Introducing commitment strategies}
We now extend the DSAIR model with strategies that can bilaterally commit to safety compliant behavior: Before an interaction, participants can commit to play SAFE in each round. The commitment stands when all parties agree. 
The players can refuse to commit, preferring to proceed without being pushed into the safe direction and being able to take risks. Those that committed but later select the UNSAFE action are potentially subject to sanctioning. Two sanctioning scenarios are considered here: (a) peer punishment (PP), which is performed by the co-player who kept her side of the deal, and (b) institutional punishment (IP), which is performed by a third-party that is not actively participating in the race for supremacy in some domain through AI (e.g. the European Union or United Nations). Each player has the freedom of behavioural choice whereby those who do not commit will not be punished when playing UNSAFE in the DSAIR model. We call the latter behavior "honest" unsafe behavior whereas those that act unsafely after committing are referred to as "dishonest".

Sanctioning an opponent who  played UNSAFE in a previous round consists in  imposing  a  reduction $s_\beta$ on the opponent's speed  \citep{han2020Incentive}. In case of PP, the punishing player also incurs a  reduction $s_\alpha$ on her own speed. 
Committing may also be costly for all that do as they give up other choices. A commitment cost $\epsilon$ (per round) is thus introduced for those performing this pre-play action. 
\begin{figure*}
\centering
\includegraphics[width=\linewidth]{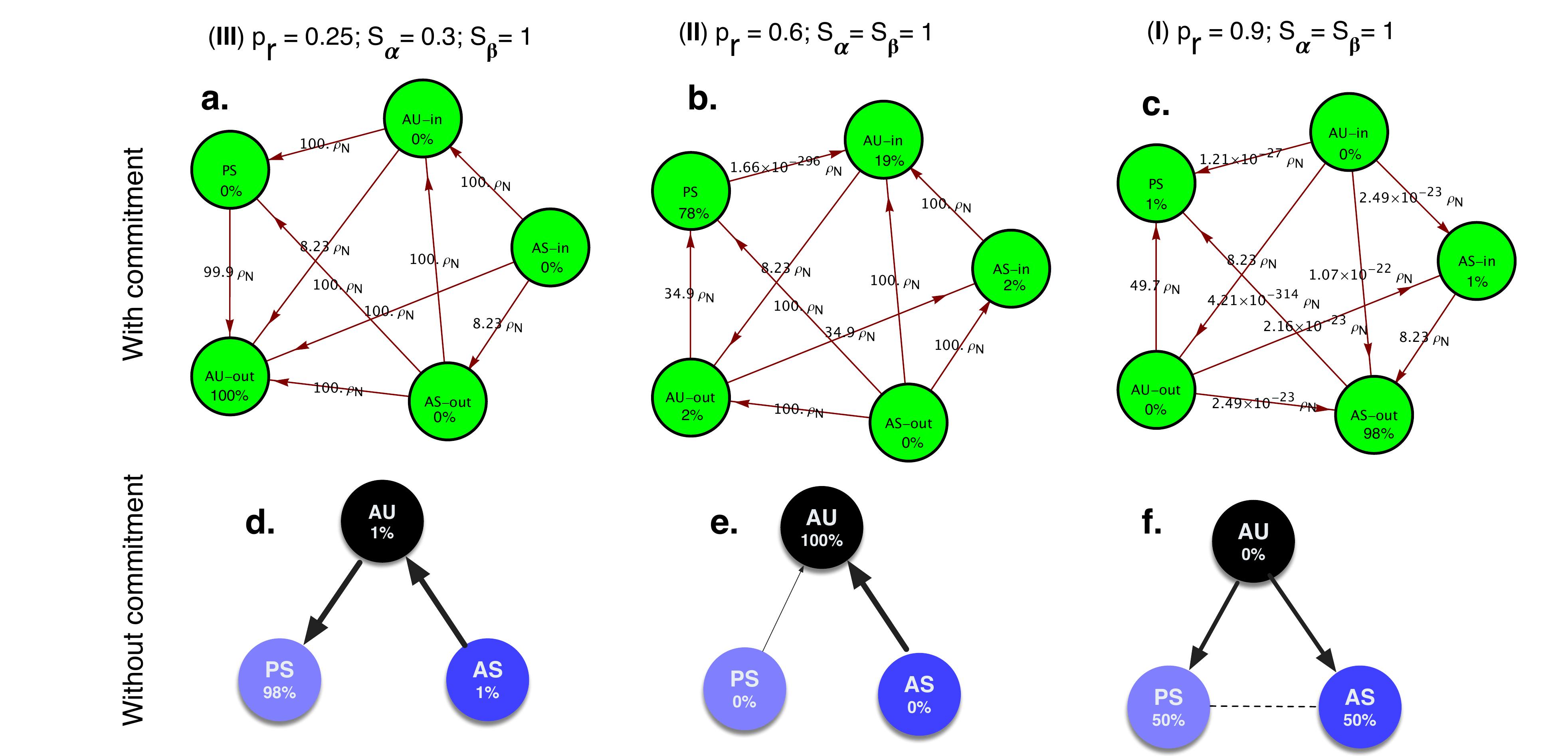}
\caption{\textbf{Transitions and stationary distributions when agreement is present (top row) against  when it is absent (bottom row), for three regions}. For clarity, only stronger transitions (than the ones in the opposite directions) are shown; no transition either way means neutral.  The choice of $s_\alpha$ and $s_\beta$ values were chosen to illustrate the main difference  between with vs without commitment scenarios, in the three zones.   %In region (III) (first column, low risk, where UNSAFE or innovation is wanted), low-cost highly efficient punishment can lead to significant reduction of innovation since PS dominates AU in absence of agreement (panel d). Addition of AU-out provides an escape as this strategy is not punished without an agreement, thereby dominating  PS (panel a).  In region (II) (dilemma zone, safety is wanted), when punishment is not highly efficient, there is a large amount of unwanted AU in the system (panel e). As AU-out is dominated by PS since PS plays UNSAFE in absence of an agreement, it leads to lower unsafe behaviour (compared panels b and e). In region (III), addition AU-out does not change the desired outcome of the dominance of safety compliance. 
Parameters: $b = 4$, $c = 1$, $s = 1.5$, $W = 100$, $B = 10^4$, $\beta = 1$,  $Z = 100$. 
}
\label{fig:panel_markov_reward_better_punishment}
\end{figure*}
With the possibility of joining or not a commitment to behave safely and sanctioning dishonest unsafe behaviour, one can now  define the possible strategies.  AS and AU (as defined above) can either commit to safe actions, and furthermore, when involved in a commitment, decide whether to punish a dishonest  co-player. If no commitment can be made, the player will select the UNSAFE action\footnote{We also consider the version where these players unconditionally play SAFE regardless of the commitment. The safety outcomes in all regions are similar, just the strategies' dominance is slightly different. Results are provided in Appendix (see Figures A3 and A4).}. The choices listed before lead to five strategies: 
\begin{enumerate} 
	\item  \textbf{AS-in}: willing to commit,  plays SAFE when commitment is in place and UNSAFE otherwise, but does not use (costly) punishment. This strategy can be considered  a second-order free-rider on the punishment effort of others;  
	\item \textbf{AS-out}: does not commit, but always selects the SAFE action; 
	\item  \textbf{AU-in}: claiming to commit to the SAFE action, but always plays UNSAFE in the interaction. This strategy makes a  commitment, trying to exploit safe players who only want to interact under an agreement; 
	\item \textbf{AU-out}: does not commit and plays  UNSAFE in the interaction. This strategy wants to freely take risk or innovate without any repercussions otherwise imposed by a commitment to follow SAFE actions; 
	\item  \textbf{PS}: willing to commit and plays SAFE when the other player also commits; plays UNSAFE otherwise,  and also punishes an UNSAFE action of a co-player that committed with her. This strategy is only present  in the case of PP, and is not present in the IP sanctioning model. 
\end{enumerate}

\subsection{Evolutionary Dynamics of Finite Populations}
%TO BRIEFLY DESCRIBE EGT FOR FINITE POPULATIONS 

Herein are adopted the methods of EGT for finite populations \citep{key:Sigmund_selfishnes,traulsen2006,hindersin2019computation}, whether in the analytical or numerical results obtained here.   In such settings,  the payoffs of players stand for their social \emph{success} or \emph{fitness}, and  the evolutionary dynamics is shaped by social learning, in accordance to which  the players that are most successful  will tend more often to be copied  by other players. The so-called pairwise rule of comparison is utilised to model social learning      \citep{traulsen2006},  which ensures   a player $A$ with fitness $f_A$ resorts to adopt the strategy of player $B$ with fitness $f_B$ with a probability {established} by the Fermi function, 
$P_{A,B}=\left(1 + e^{-\beta(f_B-f_A)}\right)^{-1}$,  
where  the intensity of selection is conveniently described by $\beta$. % ($\beta=0$ represents neutral drift while $\beta \rightarrow \infty$ represents increasingly deterministic selection). 
In a population wherein several strategies are in co-presence, their long-term frequency can  be  computed simply by calculating  the stationary distribution of a Markov chain the states of which represent each  strategy. %Details of this calculation can be found e.g. in \citep{key:Sigmund_selfishnes}.  
Absent {behavioural} exploration  or mutations, the end states of evolution are inevitably monomorphic.    {Meaning whenever}
 such a state is reached, escape by imitation is impossible. Hence, we presume further that, given some mutation probability,  each agent may     {freely explore its behavioural space (which consists of the two actions, UNSAFE  and SAFE, in our case), by randomly adopting an action as a result of mutation}.  At the limit of {a} small  {probability} of mutating,    {the population is comprised  of at most one of two strategies at any time. Accordingly,  the social dynamics is describable utilising a Markov Chain, in which each state } represents a monomorphic population whose transition probabilities to another state are expressed by the fixation probability of one single mutant \citep{key:imhof2005,key:novaknature2004}.    {The  Markov Chain's stationary distribution depicts the average time }  the whole population spends at each monomorphic end state (see the examples  in Figure \ref{fig:panel_markov_reward_better_punishment} for illustration). 

 Let $\pi_{X,Y}$ denote the payoff some strategist $X$ gathers from a pairwise interaction with some strategist $Y$ (as defined in the payoff matrix). Assume there exist two strategies at most in the population, for example $k$ agents using  strategy A ($0 \leq k \leq Z$)  and $(Z-k)$ agents using instead strategy B. Hence, the (average) payoff of agents  using  A and B can be formulated, respectively, as
\begin{equation} 
\label{eq:PayoffA}
\begin{split} 
\Pi_A(k) &=\frac{(k-1)\pi_{A,A} + (Z-k)\pi_{A,B}}{Z-1},\\
\Pi_B(k) &=\frac{k\pi_{B,A} + (Z-k-1)\pi_{B,B}}{Z-1}.
\end{split}
\end{equation} 
As a result, at each step in time, the probability    {of changing by $\pm$1 of the number of  $k$ agents using strategy A} is specified as \citep{traulsen2006} 
\begin{equation} 
T^{\pm}(k) = \frac{Z-k}{Z} \frac{k}{Z} \left[1 + e^{\mp\beta[\Pi_A(k) - \Pi_B(k)]}\right]^{-1}.
\end{equation}
The fixation probability of a single mutant {adopting} strategy A,in a population $(Z-1)$ of  agents    {adopting} B, is defined by \citep{traulsen2006,key:novaknature2004}
\begin{equation} 
\label{eq:fixprob} 
\rho_{B,A} = \left(1 + \sum_{i = 1}^{Z-1} \prod_{j = 1}^i \frac{T^-(j)}{T^+(j)}\right)^{-1}.
\end{equation} 
%In the limit of neutral selection (i.e. $\beta = 0$), $\rho_{B,A}$ equals the  inverse of population size, $1/N$. 
When considering a set  $\{1,...,s\}$ of different strategies, such probabilities of fixation  define the Markov Chain transition matrix $M = \{T_{ij}\}_{i,j = 1}^s$, with $T_{ij, j \neq i} = \rho_{ji}/(s-1)$ and  $T_{ii} = 1 - \sum^{s}_{j = 1, j \neq i} T_{ij}$. The normalized eigenvector    {of the transposed matrix of $M$ associated with eigenvalue 1} produces the above defined  stationary distribution  \citep{key:imhof2005}, depicting the relative time the population stays adopting each of the strategies. 

\subsection{Risk-dominance} 

A major {standpoint} of comparison of  two strategies A and B is in which direction the transition is more probable or stronger, the one of some B mutant  fixating in a population of agents that employ A, $\rho_{A,B}$,  or that of a mutant A fixating in the population of agents that employ B, $\rho_{B,A}$.    {At the limit, for a large enough population size (i.e. a large $Z$), the condition  simplifies to} \citep{key:Sigmund_selfishnes}  
%\begin{equation} 
%(N-2)\pi_{A,A} + N\pi_{A,B} > (N-2)\pi_{B,A} + N\pi_{B,B}
%\end{equation} 
%which,is simplified to 
\begin{equation} 
\label{eq:risk_dominance Equation}
\pi_{A,A} + \pi_{A,B} > \pi_{B,A} +  \pi_{B,B}.
\end{equation}

\begin{figure*}[h!]
\centering
\includegraphics[width=\linewidth]{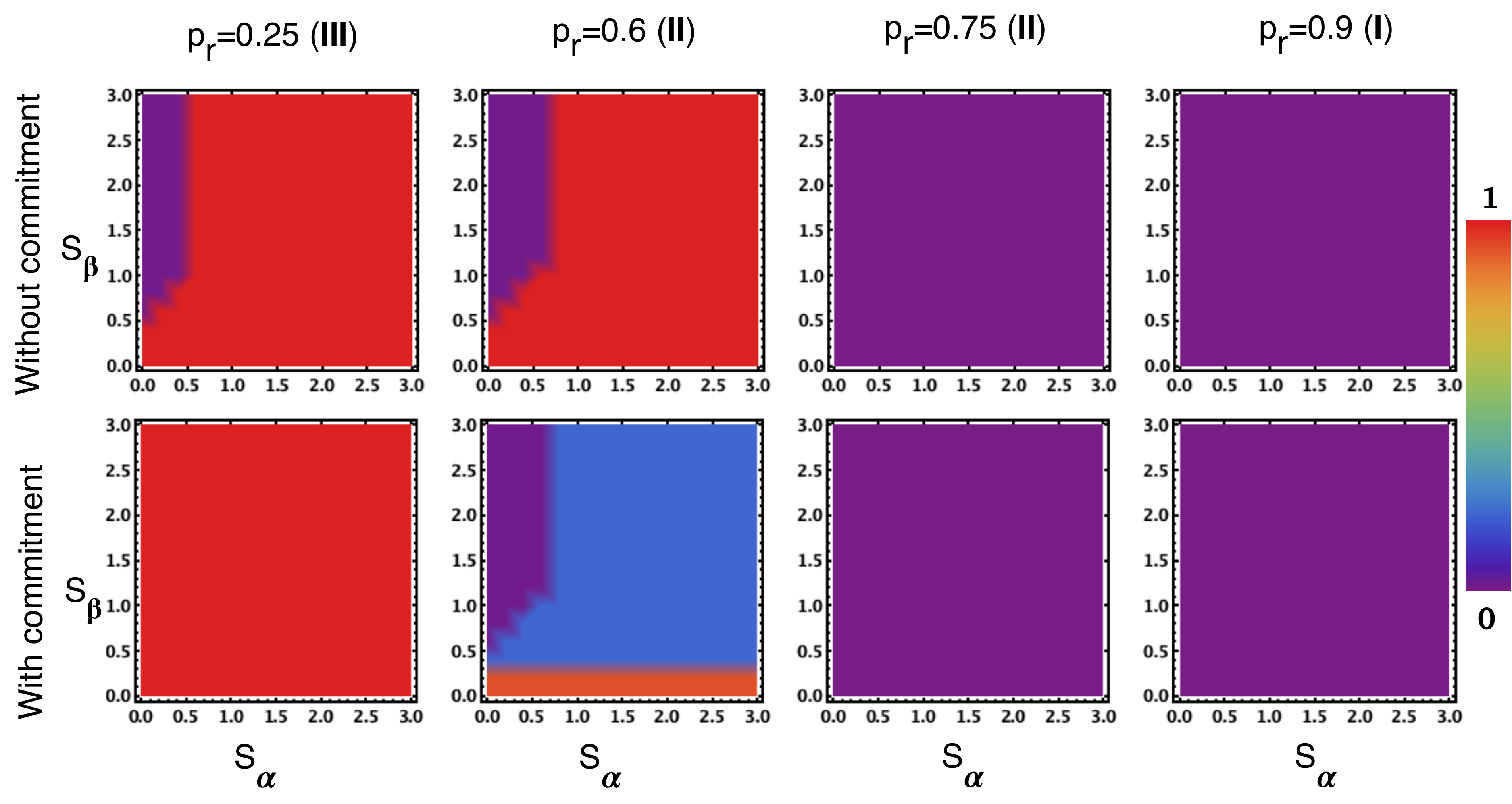} 
\caption{\textbf{Frequency of unsafe  in all regions for varying $s_\alpha$ and $s_\beta$, when commitment is absent (top row) vs when it is  in use (bottom row)}.  %In region (III), over-regulation occurs in the former case when $s_\alpha$ is sufficiently small and $s_\beta$ is large, which is not the case in the latter one. In  region (II) with the lower range of $p_r$ (second column, $p_r = 0.6$), unsafe frequency is lower in the latter in for most $s_\alpha$ and $s_\beta$. In the higher range  of region (II) and region (III) (third and fourth columns, with $p_r = 0.75$ and $p_r = 0.9$), desired safe behaviour is dominant  in both cases.  
Parameters:  $b = 4$, $c = 1$, $s = 1.5$, $W = 100$, $B = 10^4$, $\beta = 1$,  $Z = 100$. 
}
\label{fig:incentives_effect}
\end{figure*}

\section{Results}
We first focus on analysing the system of self-enforcing bilateral commitment (PP)  then showing the results  of bilateral commitment complemented with institutional enforcement (IP). 

\subsection{Self-enforcing bilateral commitments}
In Figure \ref{fig:agreementvsnoagreement} (top row), we show the stationary distributions of the five strategies for varying $p_r$ across the three zones \textbf{I}, \textbf{II} and \textbf{III}. 
We show results for  two important scenarios: when efficient punishment can be made for a small cost (\textit{left column:} $s_\alpha = 0.3$, $s_\beta = 1$) and when punishment is not highly efficient  (\textit{right column:}, $s_\alpha = 1$, $s_\beta = 1$). Punishment is considered efficient when its effect  ($s_\alpha$) on the development speed of the player  performing the punishment is significantly smaller  than the  effect ($s_\beta$) it has on the player undergoing the punishment. 

In both cases, AU-out dominates when $p_r$ is small (zone \textbf{III}), PS dominates when it is intermediate (first part of zone \textbf{II}),  while AS-out dominates when it is large (part of zone \textbf{II} and zone \textbf{I}). It is important to notice that these results reflect the most desirable outcomes for all three zones: risky innovation in \textbf{III} and safety compliance in \textbf{I} and \textbf{II} (see the black line in the bottom row of Figure \ref{fig:agreementvsnoagreement}). Moreover,  a more efficient punishment leads to better safety outcome in zone \textbf{II} (see panels c and d). It is worth noting that inefficient punishment may result in the presence of AU-in strategists, in other words cheaters. It is thus important in this self-organised commitment solution that participants have an effective mechanism to punish commitment violators. One solution could be directly affecting the player's publicly available reputation. 

To clarify the benefit of voluntary bilateral commitment, Figure \ref{fig:agreementvsnoagreement} (bottom row) compares the overall frequency of  unsafe/risk-taking when commitment is possible (black line) vs when it is absent (red line). In the absence of bilateral commitments, 
over-regulation  occurs, i.e. safety compliance is abundant but not desired for a large part of zone \textbf{III} when efficient punishment can be carried out for a small cost (panel c), while a safety dilemma occurs (i.e. safety compliance is desired but infrequent) for a large part of zone \textbf{II} when sanctioning is not highly efficient. 

These observations can be better understood by examining
the transitions and stationary distributions in Figure \ref{fig:panel_markov_reward_better_punishment}.
In region \textbf{III} (first column, low risk, where risky innovation is desired), low-cost highly-efficient punishment can lead to significant reduction of innovation since PS dominates AU in absence of bilateral commitments (panel d). Addition of AU-out provides an escape as this strategy is not punished since it never commits to safe course of actions, thereby dominating PS (panel a).  In region \textbf{II} (dilemma zone, safety is wanted but needs to be enforced), when punishment is not highly efficient, there is a large amount of unwanted AU in the population (panel e). As AU-out is dominated by PS, since PS plays also UNSAFE when not having a commitment partner, it leads to a lower frequency of  unsafe behaviour (comparing panels b and e). In region \textbf{III}, addition AU-out does not change the desired outcome of the dominance of safety compliance.

In Figure \ref{fig:incentives_effect}, we show that these remarkable observations regarding voluntary bilateral commitment are robust for different regimes of punishment effectiveness. In particular, when comparing the unsafe frequency with commitments against when it is absent for varying  $s_\alpha$ and $s_\beta$. 
In region \textbf{III}, over-regulation occurs when there is only punishment whenever $s_\alpha$ is sufficiently small and $s_\beta$ is large (purple area, top row, first column), but when an agreement is in place that is not the case  (bottom row, first column). In  region \textbf{II} with the lower range of $p_r$ (second column, $p_r = 0.6$), unsafe frequency is lower in the latter in for most $s_\alpha$ and $s_\beta$. In the higher range  of region \textbf{II} and region \textbf{III} (third and fourth columns, with $p_r = 0.75$ and $p_r = 0.9$), desired safe behaviour is dominant  in both cases.
 
It's noteworthy that all results are robust also for other regimes of weaker or stronger selection intensities, i.e.  $\beta = 0.1 $  or $\beta = 10$ (see e.g. Figures \ref{fig:vary_pr-otherbetas} and A2 in Appendix).

\subsection{Institutionally governed bilateral commitments}
When assuming that it is an institution that governs the dynamics of the race to supremacy through AI in some domain, punishing unsafe behaviour if a bilateral commitment is violated. Similarly to PP, IP reduces the development speed of a participant that selected the UNSAFE action by $s_\beta$.  

On the one hand, when no bilateral commitments can be made, the population consists of two strategies, AS and AU, where AU's speed will be $s - s_\beta$ when being punished (by the institution). On the other hand, when  voluntary commitments are possible, there will only be the four strategies,  AS-in, AU-in, AS-out, and AU-out. There is no PS strategy as there is no peer sanctioning, which implies also no parameter $s_\alpha$  given that the institution is not part of the population\footnote{Future work might look at how to minimise the cost and efforts from the institution while  ensuring a desired level of safety compliance  \citep{han2018cost,Hanijcai2018,chen2015first,wang2019exploring,couto2020governance}}.  
Concretely, the institution can only sanction AU-in when both players committed. In that case, its speed will be reduced, from $s$ to $s -s_\beta$. AU-out as before will be free of any sanction. 

\begin{figure}[h!]
\centering
\includegraphics[width=\linewidth]{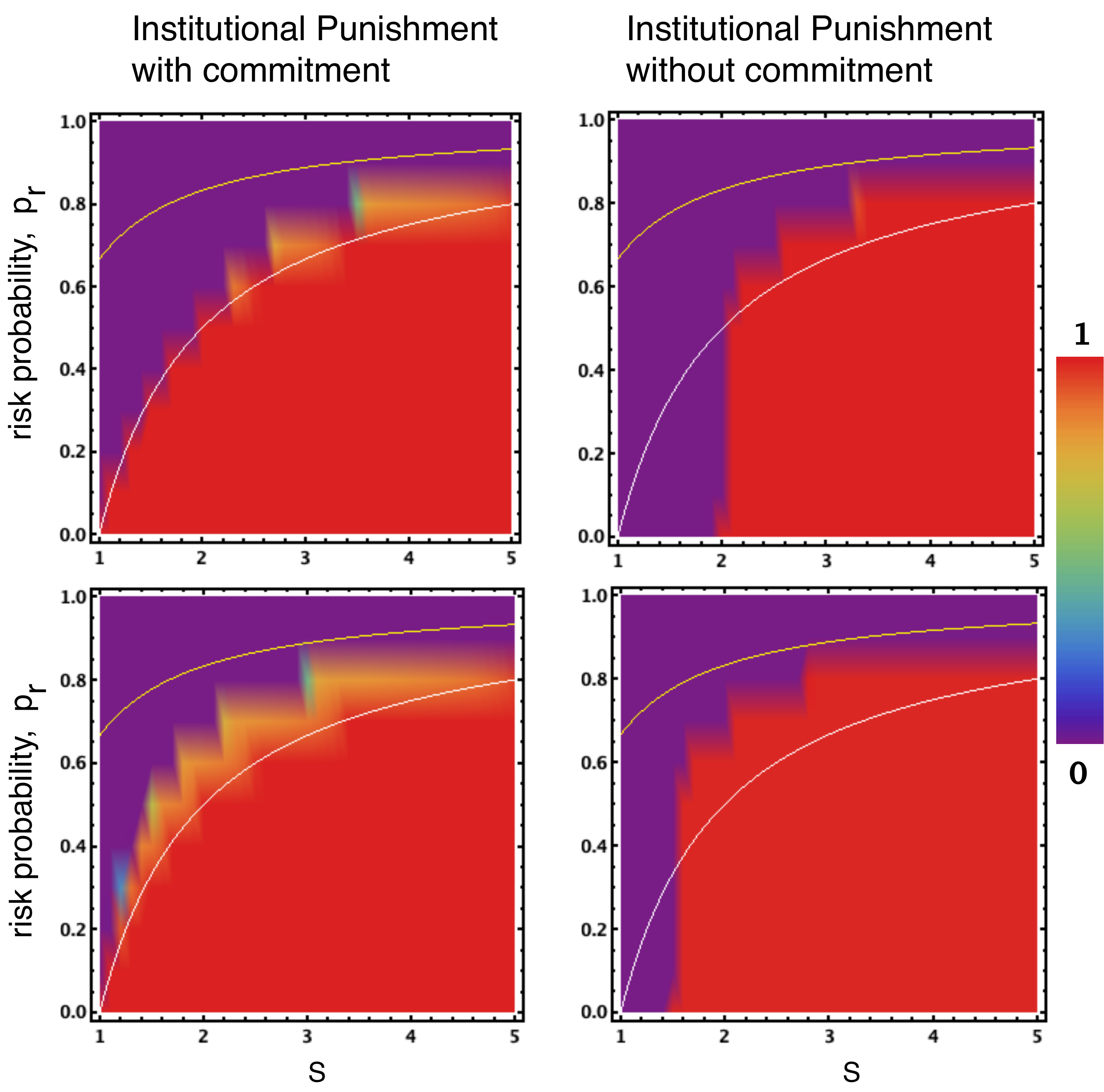} 
\caption{\textbf{Institutional punishment, with (left column) or without a safety agreement (right column).} We report the frequency of unsafe behaviour (UNSAFE) in a population of AU and AS in the latter case and a population of AS-in, AU-in, AS-out, AU-out, in the former case. An institutional punishment reduces the speed of AU in the latter and AU-in (when an agreement is formed) in the former, by $s_\beta = 1$ (\textbf{top row}) and by $s_\beta = 0.5$ (\textbf{bottom row}). The lines within the figures are as in Figure 1.   Parameters:  $b = 4$, $c = 1$, $W = 100$, $B = 10^4$, $\beta = 1$,  $Z = 100$. 
}
\label{fig:institutional_punishment_panel}
\end{figure}

In Figure \ref{fig:institutional_punishment_panel}, we compare the effect of institutional sanctioning with and without the option of having prior voluntary bilateral commitments between the two players. The outcomes are similar to the case of PP discussed in the previous section. In particular, one can observe that the presence of commitments can significantly improve safety outcome in region \textbf{II} without being detrimental to the desired behaviour and thus social outcome in the other two regions. Without commitments, institutional punishment leads again to over-regulation of unsafe/innovation in region  \textbf{III}, which, as was mentioned before, is not a desired outcome. Moreover, with a smaller  $s_\beta$ (compare  top and bottom rows of the figure), the outcomes remain  similar, with a  lower level of  over-regulation but weaker results for region \textbf{II}. 

In short, we observe that freedom of choice between bilateral commitments to safety regulations (which can be self-enforcing or institutionally enforced) and unilateral risky/unsafe endeavours provides an efficient solution to ensure desirable outcomes in all scenarios of the DSAIR model. First, it alleviates the problem of over-regulation when the risk is low (which happens when punishment can be very efficient), while at the same time it  reduces the frequency of unsafe behaviour in the dilemma zone (region \textbf{II}). Furthermore,  the desirable safety compliance outcomes in region \textbf{I} remain unaffected.

\section*{Discussion and conclusions}
 
This paper proposes and analyses, using a multi-agent and population dynamics modelling approach,  a novel solution that promotes desirable safety compliance in a technology development race, while at the same time avoiding stifling beneficial innovation due to over-regulation. We base our study on  a previously proposed EGT model  \citep{han2019modelling} that describes  the dynamics of a competition between safety and risk-taking (unsafe) behaviours, within  an (alleged) race for  supremacy through AI in a marketable domain.
We show that, by allowing race participants the freedom of choosing to enter or not in bilateral commitments to act safely and avoid risks, accepting thus to be sanctioned in case of misbehavior,  high levels of the most beneficial behaviour as a whole, are achieved in all regions of the parameter space.

This system of voluntary bilateral commitments, either depending on sanctioning actions of peers or by an institution, provides a mechanism to overcome the problems associated with over-regulation, which  might occur whenever risk taking behaviour, which may be perceived as unsafe, is unconditionally penalised without taking into account the true risk level (in relation to the  cost and benefit of the safety compliance and of the opting for risky behaviour), as shown in  \citep{han2020Incentive}.
On the one hand, allowing for participants to explicitly commit to safety precautions in their development process ensures higher levels of safety compliance when that is desired, because then unsafe/risk-taking strategies become either clearly identified  (AU-out) and suitably coped with, or punished through the binding commitment (AU-in).  On the other hand, the commitment option enables the (honest) risk-taking behaviour (namely, AU-out) to prevail  whenever it is collectively preferred, since then it is not punished, as a result of not joining the commitment in the first place.

There have been several theoretical modelling studies based on EGT,  showing the benefit of establishing  prior commitments or agreements for promoting the evolution of certain positive behaviours, such as  cooperation and coordination in a population of self-regarding agents, see e.g.    \citep{han2013good,sasaki2015commitment,HanJaamas2016,bianca2021coordination_agreement}. Behavioral  experiments with human subjects have also been performed showing the promoting role of commitments for cooperative behaviours, see e.g.   \citep{chen1994effects,cherry2013enforcing}. In all such contexts, it is well-defined what is the correct (positive) behaviour most beneficial collectively and thus the one that should be promoted. 
In our case that no longer holds, since whether or not a  behaviour is the most beneficial collectively depends on the behavioural region in which the race occurs at the time. This region is not well defined, especially when data is not abundantly available.  For example, we might need to wait for a technology to be audited and even largely adopted by users to know the level of risk associated with a particular technology (but then it might be too late already to provide pertinent regulations). Such level of uncertainty can also lead to other externalities, such as the emergence of strongly polarized positions \citep{domingos2020}, an effect not yet characterized by our model.

Interestingly, we show here that establishing the possibility of a voluntary bilateral commitments is as well highly effective in promoting collectively desired/preferred behaviour in all cases. 
Commitment to cooperative actions in social dilemmas work because they allow cooperators to avoid free-riding strategies, for the case of an AI race, because they permit players the freedom to follow their preferred course of development without being punished (as long as they are being honest). One can also see it as a particular form of binding signal or pledge. The results of this work are in line with  Hadfield's arguments \citep{hadfield2017rules} in the general context of technology globalisation; that is, legal contracts and  institutions should allow freedom (via coordination) instead of exacting it from individuals in a population. For the sake of clarity, here we illustrate this idea with a pairwise race model; however, as shown in \citep{han2019modelling}, this model can be easily extended to N-player interactions, with a concomitant increase in complexity, yet leading to the same qualitative messages. Future work may explicitly consider N-player commitments and more complex communication and bottom-up dynamics typical of this type of agreements, see e.g. \citep{HanJaamas2016,Han2017AAAI,vasconcelos2013bottom}. 

Last but not least, in \citep{o2020windfall} the authors propose the so-called  Windfall Clause for mediating the tension in AI competition.  It is proposed there to arrange prior commitments or agreements from the race participants (e.g., companies, governments) for sharing a large part of the windfall benefit (i.e. the large prize $B$ in our model) with society. As discussed in that paper, it requires significant legal infrastructures to be put in place to enable  such a mechanism. Our approach does not so require because legal contracts can be used to ensure  compliance with prior safety agreements.

  It is noteworthy that, although we focus  in this paper on an AI development race, the model proposed can be  more generally applicable to other kinds of long-term  situations of competition, such as the development of technological innovation  {and its racing for patents} in which there is a significant advantage (i.e. a large $B$) to be gotten in being one of the first ever to reach an important target \citep{denicolo2010winner,campart2014technological,lemley2012myth}.
Other important domains involve pharmaceutical and vaccines  development race, where companies might attempt to cut a few corners by not  following strictly the safe clinical trial protocols, with the view to be the first ones to develop and put on the market some pharmaceutical product, in order to reap the highest possible share of benefit in the market \citep{abbott2009global,burrell2020covid}. However,  certain aspects or factors of the current model might need to be revised.
For example,  the large profit  $B$ for a vaccine development race would be more difficult to achieve than in the case of AI, since a developed vaccine needs to be approved by suitable authorities before users have trust in it and use it.
Vaccine developers only can generate the profit $B$ if they could show evidence of the safety of the vaccine and that they have closely followed  all the safety procedures.
In case of AI, the winning developers  can simply deploy the technologies and get the profit B (in most cases). One can expand the current model to capture also the vaccine race (and also generalised the current AI race model) by for example having a new parameter to capture when $B$ can be generated for the race winners, which depends on the nature of the race (e.g. vaccine vs AI) and also the frequency of safety compliance in the past. our future works will address  these issues.

%Besides its tremendous  economic advantage, a winner of a vaccine race, say for Covid-19 prevention, can also glean significant reputational  and political influence \citep{}. 

\section{Acknowledgements}
T.A.H., L.M.P., T.L. and T.C., are supported by Future of Life Institute grant RFP2-154.  L.M.P. acknowledges support from FCT/MEC NOVA LINCS PEst UID/CEC/04516/2019. F.C.S. acknowledges support from FCT Portugal (grants PTDC/EEI-SII/5081/2014, PTDC/MAT/STA/3358/2014). T.L. acknowledges support by the FuturICT2.0 (www.futurict2.eu) project funded by the FLAG-ERA JCT 2016. T.A.H. is also supported by a Leverhulme Research Fellowship (RF-2020-603/9).

%\newpage

 \renewcommand{\thefigure}{A\arabic{figure}}
 \renewcommand{\thetable}{A\arabic{table}}
 \setcounter{figure}{0}

\newpage
\section{Appendix} 
\subsection{Effect of larger intensity of selection} See Figures A1 and A2. 
\subsection{Other scenarios of modelling}
See Figures A3 and A4. 

\begin{figure*}[h!]
\centering
\includegraphics[width=0.8\linewidth]{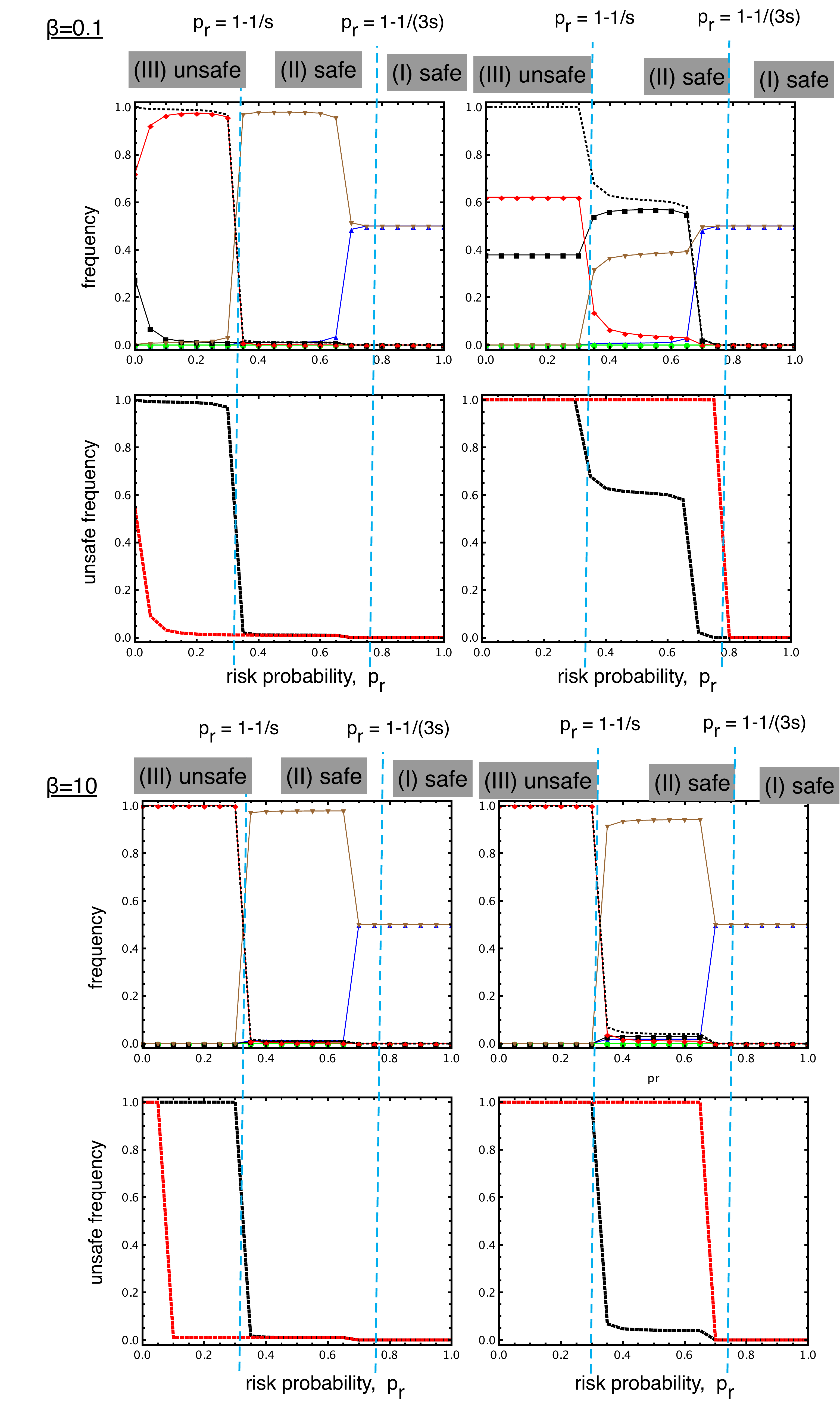} 
\caption{\textbf{Same as  Figure 2,  with other values of  $\beta$}.% Behavioural dominance of in different zones for varying $p_r$, in presence of prior agreement  (top row: panels a, b) and comparison of its overall unsafe behaviour against when agreement is absent  (bottom row: panels c, d).  %Parameters:  $b = 4$, $c = 1$, $s = 1.5$, $W = 100$, $B = 10^4$, $Z = 100$. 
}
\label{fig:vary_pr-otherbetas}
\end{figure*}

\begin{figure*}[h!]
\centering
\includegraphics[width=\linewidth]{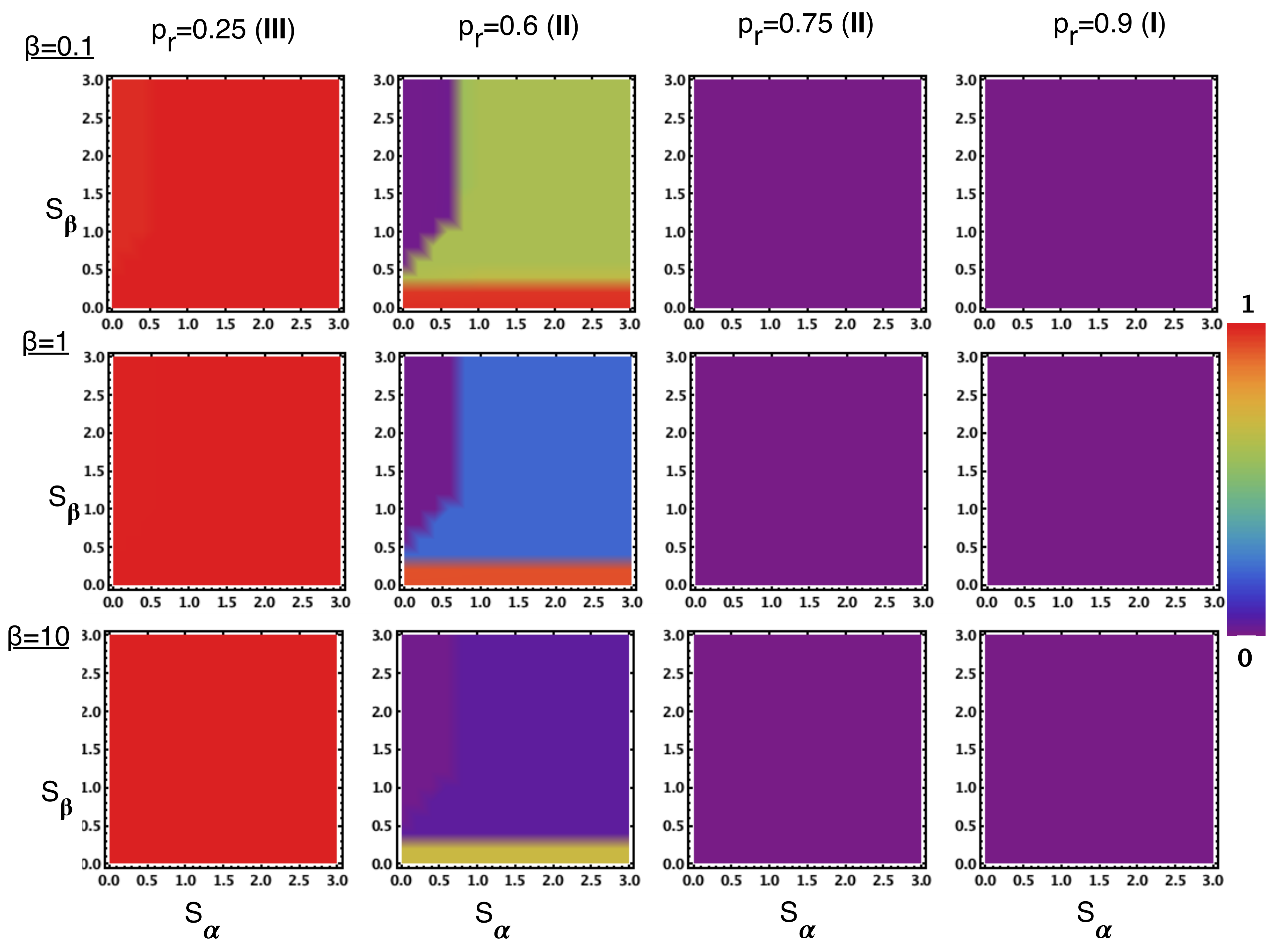} 
\caption{Frequency of unsafe  in all regions for varying $s_\alpha$ and $s_\beta$, when safety agreement is  in use, for different values of intensities of selection $\beta$.  Similar observations as in the main text.   Parameters:  $b = 4$, $c = 1$, $s = 1.5$, $W = 100$, $B = 10^4$,   $Z = 100$. 
}
\label{fig:panel_agreement_countour_unsafe_different-betas}
\end{figure*}

\begin{figure*}[h!]
\centering
\includegraphics[width=0.9\linewidth]{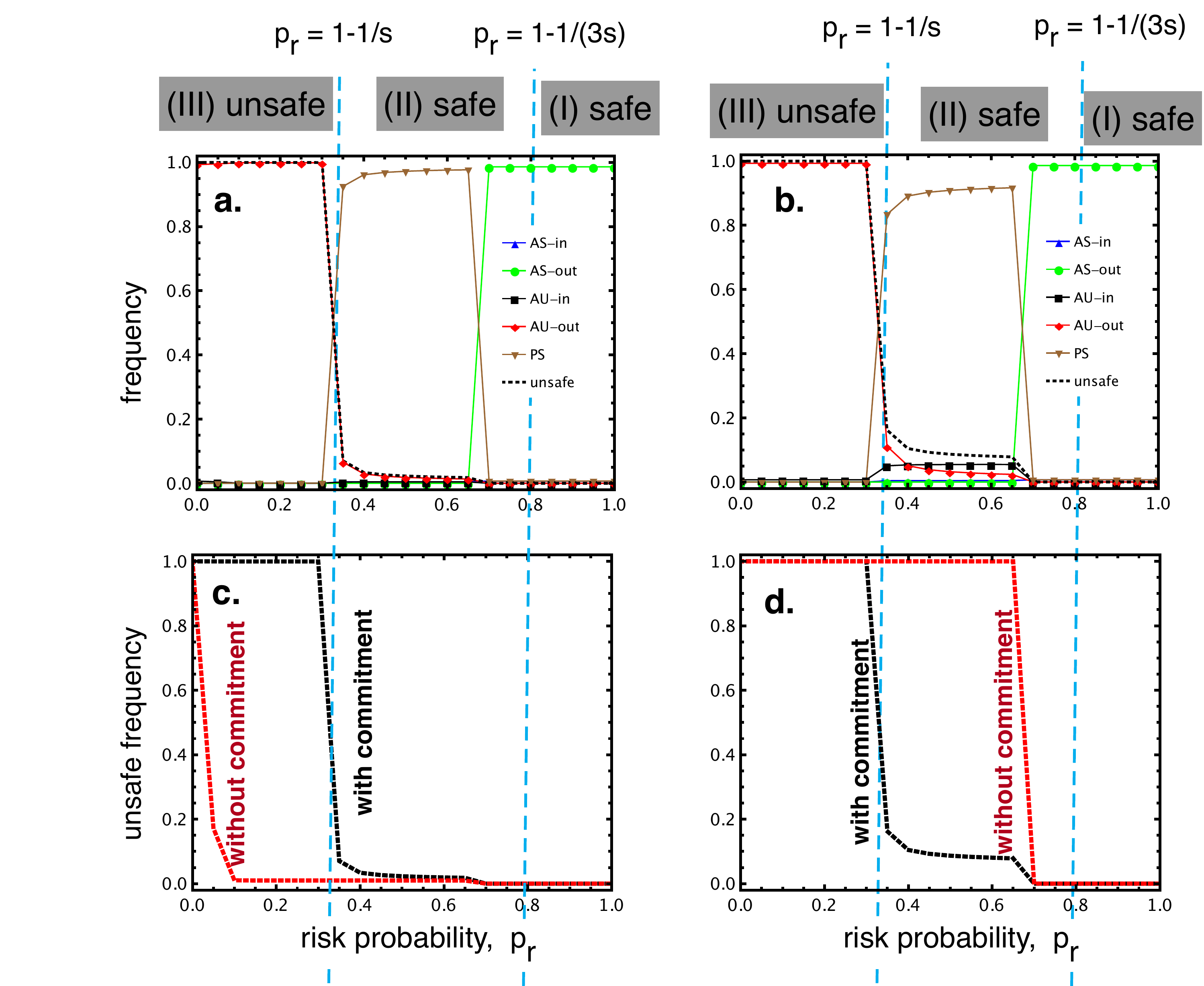} 
\caption{When AU-in  always plays SAFE in an interaction. Behavioural dominance of in different zones for varying $p_r$, in presence of prior commitment  (top row: panels a, b) and comparison of its overall unsafe behaviour against when commitment is absent  (bottom row: panels c, d). We show results for  two important scenarios: when efficient punishment can be made for a small cost (\textit{left column:} $s_\alpha = 0.3$, $s_\beta = 1$) and when punishment is not highly efficient  (\textit{right column:}, $s_\alpha = 1$, $s_\beta = 1$). 
In both cases, AU-out dominates when $p_r$ is small (zone \textbf{III}), PS dominates when it is intermediate (first part of zone \textbf{II}),  while PS and AU-in  together dominate when it is large (part of zone \textbf{II} and zone \textbf{I}). That results in the fact that when a commitment is present desirable outcomes are achieved in all three zones: unsafe/innovation in \textbf{III} and safe in \textbf{I} and \textbf{II}, see bottom row. In absence of a commitment, 
over-regulation  occurs (i.e. safe is abundant but not desired) for a large part of zone \textbf{I} when efficient punishment can be done for a small cost (panel c), while safety dilemma occurs (i.e. safe is desired but infrequent) for a large part of zone \textbf{II} when punishment is not highly efficient.   Parameters:  $b = 4$, $c = 1$, $s = 1.5$, $W = 100$, $B = 10^4$, $\beta = 1$,  $Z = 100$. 
}
\label{fig:incentives_effect-SI}
\end{figure*}

\begin{figure*}
\centering
\includegraphics[width=\linewidth]{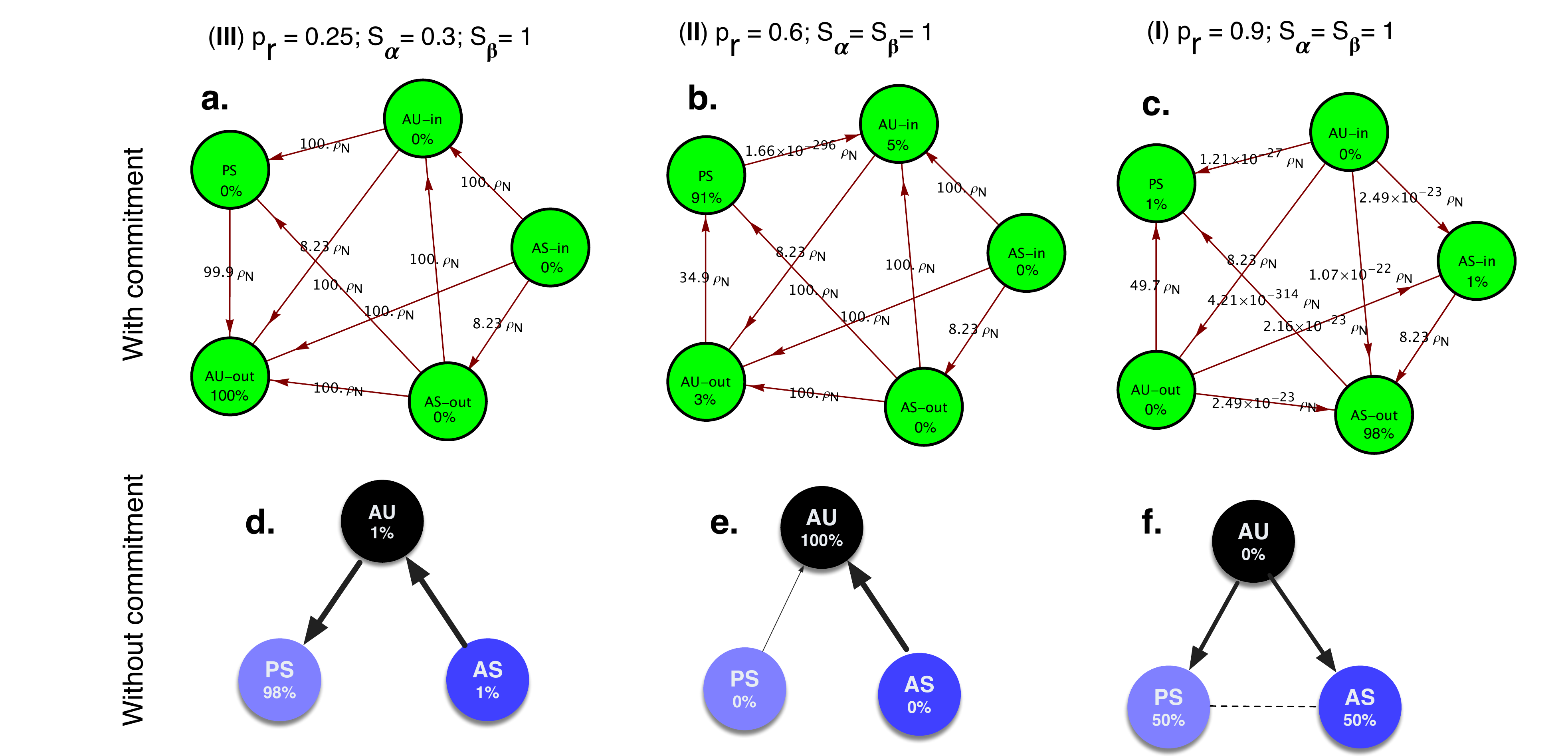}
\caption{When AU-in  always play SAFE in an interaction. Transitions and stationary distributions when a commitment is present (top row) against  when it is absent (bottom row), for three regions (Only stronger transitions are shown; no transition either way means neutral). In region \textbf{III} (first column, low risk, where UNSAFE or innovation is wanted), low-cost highly efficient punishment can lead to significant reduction of innovation since PS dominates AU in absence of a commitment (panel d). Addition of AU-out provides an escape as this strategy is not punished, thereby dominating  PS (panel a).  In region \textbf{II} (dilemma zone, safety is wanted), when punishment is not highly efficient, there is still some amount of unwanted AU in the system. As AU-out is dominated by PS since PS plays UNSAFE in absence of an commitment, it leads to lower unsafe behaviour. In region \textbf{III}, addition AU-out does not change the desired outcome of safety.  Parameters: $b = 4$, $c = 1$, $s = 1.5$, $W = 100$, $B = 10^4$, $\beta = 1$,  $Z = 100$. 
}
\label{fig:panel_markov_reward_better_punishment-SI}
\end{figure*}

\clearpage
\bibliographystyle{apalike}
\bibliography{bibliography} 
 
\end{document}